\documentclass[conference]{IEEEtran}
\IEEEoverridecommandlockouts
\usepackage{cite}
\usepackage{booktabs}
\usepackage{amsmath,amssymb,amsfonts}
\usepackage{algorithmic}
\usepackage{graphicx}
\usepackage{textcomp}
\usepackage{xcolor}
\def\BibTeX{{\rm B\kern-.05em{\sc i\kern-.025em b}\kern-.08em
    T\kern-.1667em\lower.7ex\hbox{E}\kern-.125emX}}
\begin{document}

\title{Knowledge-Driven Multi-Agent Reinforcement Learning for Computation Offloading in Cybertwin-Enabled Internet of Vehicles
}

\author{\IEEEauthorblockN{Ruijin Sun, Xiao Yang, Nan Cheng, Xiucheng Wang, Changle Li}
\IEEEauthorblockA{State Key Lab of ISN, Xidian University, Xi'an 710071, P.R. China\\}
Email: sunruijin@xidian.edu.cn; xyang\_4@stu.xidian.edu.cn; dr.nan.cheng@ieee.org; \\
xcwang\_1@stu.xidian.edu.cn; clli@mail.xidian.edu.cn
}

\maketitle

\begin{abstract}
By offloading computation-intensive tasks of vehicles to roadside units (RSUs), mobile edge computing (MEC) in the Internet of Vehicles (IoV) can relieve the onboard computation burden. However, existing model-based task offloading methods suffer from heavy computational complexity with the increase of vehicles and data-driven methods lack interpretability. To address these challenges, in this paper, we propose a knowledge-driven multi-agent reinforcement learning (KMARL) approach to reduce the latency of task offloading in cybertwin-enabled IoV. Specifically, in the considered scenario, the cybertwin serves as a communication agent for each vehicle to exchange information and make offloading decisions in the virtual space. To reduce the latency of task offloading, a KMARL approach is proposed to select the optimal offloading option for each vehicle, where graph neural networks are employed by leveraging domain knowledge concerning graph-structure communication topology and permutation invariance into neural networks. Numerical results show that our proposed KMARL yields higher rewards and demonstrates improved scalability compared with other methods, benefitting from the integration of  domain knowledge.
\end{abstract}


\section{Introduction}
The Internet of Vehicles (IoV) has received significant attention from academic and industrial communities in recent years as a fundamental technology for developing intelligent transportation systems. By seamlessly connecting people, vehicles, roads and clouds, and sharing road information and collaborative tasks among them, Iov enhances traffic efficiency and ensures driving safety. Various applications like path planning and assisted/autonomous driving are realized in the context of IoV. The deployment of computationally heavy applications in IoV, however, faces a number of difficulties, one of which is the constrained computational capacity of the vehicles. It takes a long time and uses a lot of energy to process these applications on in-vehicle terminals. To address this difficulty, mobile edge computing (MEC) has been proposed as an emerging solution \cite{ali2018mobile}, which offloads computation-intensive tasks from vehicles to roadside units (RSUs) through vehicle-to-infrastructure (V2I) communications. Compared with centralized cloud computing, MEC significantly reduces communication latency in task offloading due to the reduced distance~\cite{li2020deep}.

In MEC-enabled IoV, task offloading and resource scheduling is a hot research topic in recent years. In \cite{chen2022dynamic}, Chen \emph{et al.} proposed a joint deep reinforcement learning (DRL) algorithm with the objective of minimizing latency and energy consumption to optimize the task offloading. In \cite{zhu2022learning}, Zhu \emph{et al.} proposed a load balance scheme, where aerial relays are utilized to establish relay connections between vehicles and nearby heterogeneous edge infrastructures. In \cite{xu2023joint}, a joint computation offloading and resource allocation scheme based on the non-orthogonal multiple access architecture is proposed. Li \emph{et al.} in \cite{li2020deep} proposed a task partitioning and scheduling algorithm that determines workload allocation and execution order for tasks offloaded to edge servers. However, these DRL methods for task offloading are conducted in a centralized way, requiring a lot of information collection via wireless links and resulting in long  latency.

To move this obstacle, a cybertwin-enabled network architecture has recently been proposed in \cite{yu2019cybertwin}, where cybertwin serves as an intelligent communication agent in edge networks for its corresponding vehicle and can guide the resource allocation among vehicles in a distributed way. To be specific, cybertwin, as a virtual communication assistant, can mimic the vehicle's behavior via interaction with its physical vehicle and learning vehicle's preferences. Thus, it is usually authorized to make resource allocation decisions on behalf of the vehicle. For resource scheduling in cybertwin-enabled IoV, decision-making among vehicles is achieved by distributed interaction among their cytertwins in virtual space via wired links. Owing to this benefit, cybertwin-enabled network architecture has been applied in  \cite{9860495} and \cite{9174795} to improve the model convergence performance and reduce the offloading latency, respectively. 

Another key challenge of  existing DRL methods for task offloading is the lack of interpretability and scalability, which usually adopt deep neural networks (DNN) and convolutional neural networks.  While these architectures are suitable for image processing tasks, they do not consider the features of wireless communication networks, resulting in inefficient performance. Incorporating communication-specific domain knowledge into the architecture of neural networks is a promising  way of dealing with this issue \cite{sun2022improving}. In \cite{guo2020structure}, Guo {\it et al.} revealed that most multi-user wireless tasks have permutation invariance and then proposed a lightweight DNN with the parameter sharing scheme. In \cite{9252917}, Shen {\it et al.}  adopted graph neural networks (GNN) with permutation invariance, by embedding the graph-structured wireless network topology into neural networks, to  optimize the power allocation, which shows better scalability.

To realize distributed learning with improved scalability, in this paper, we propose a knowledge-driven DRL for task offloading in cybertwin-enabled IoV, which embeds communication topology into neural networks. Specifically, we design the cybertwin for each vehicle in its virtual space, which enables information management and information exchange among vehicles. In this architecture, we treat the digital mapping of each vehicle in virtual space as an agent and employ a centralized training and decentralized execution (CTDE) approach to train multi-agent reinforcement learning algorithms to minimize the latency. The proposed algorithm incorporates GNN into the neural network, leveraging the permutation invariance of GNN to incorporate prior knowledge into the network architecture.

\section{System Model and Problem Formulation}
\label{sec:system}

\subsection{Network Model}
\begin{figure}[htbp]
	\centering
	\includegraphics[width=.48\textwidth]{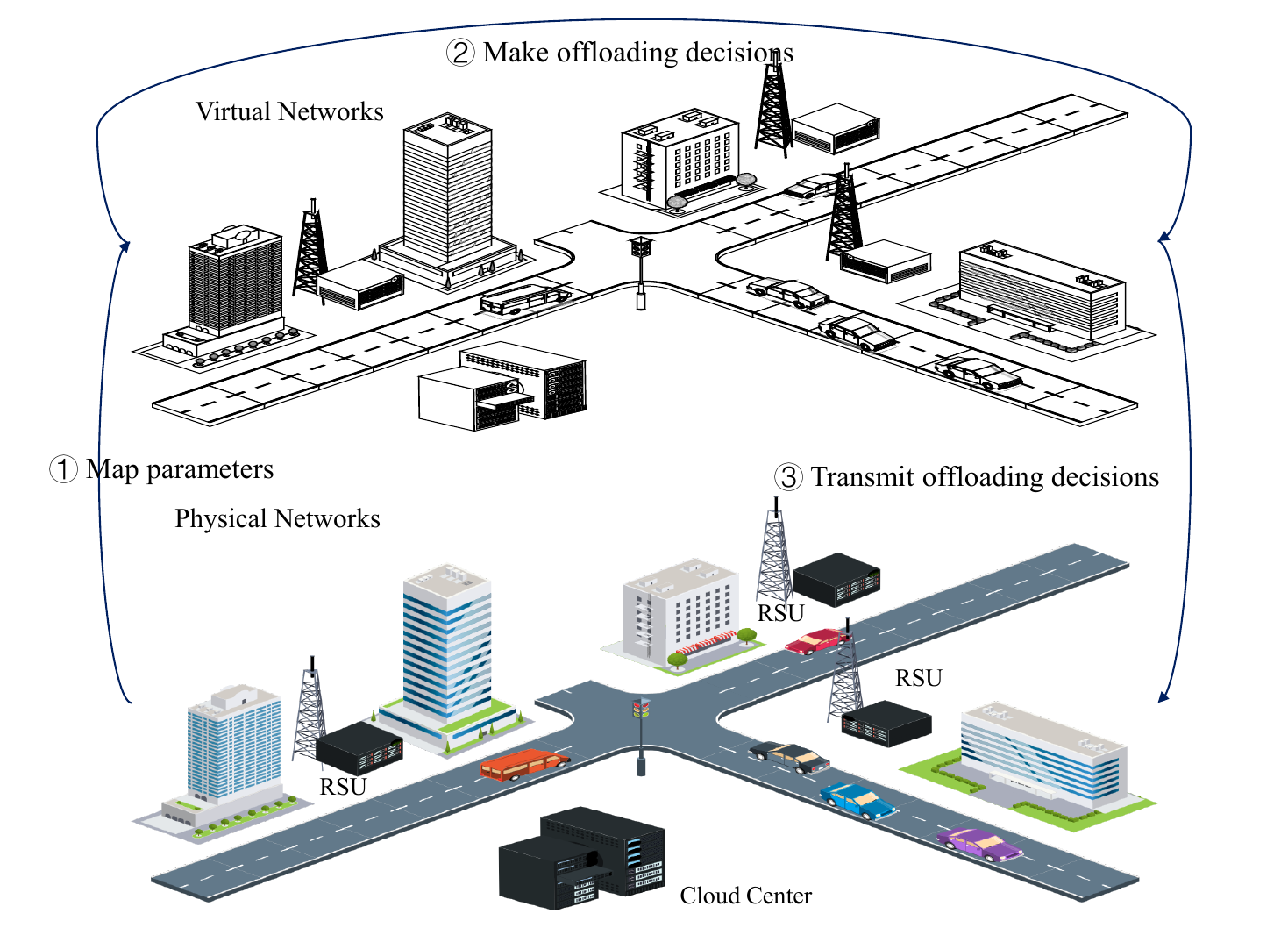} 	
	\caption{Architecture of cybertwin-enabled IoV network.}		
	\label{demo}							
\end{figure}
We consider a vehicle-edge cybertwin-enabled IoV system, which consists of multiple vehicles, multiple RSUs, and a macro base station (MBS), as shown in Fig. \ref{demo}. Each vehicle $i$ corresponds to a digital mapping $\widehat{i}$ in the virtual space. Cybertwin is deployed in a distributed manner on the RSUs, collecting real-time network state information. By mapping the parameters and executing decisions in the virtual network, cybertwin optimizes the solution and returns it to the actual network, ultimately improving traffic efficiency.

Let $\mathcal{I}=\{1,2, \ldots, I\}$ be the index set of vehicles, and $\mathcal{R}=\{1,2, \ldots, R\}$ be the index set of RSUs. We consider the IoV system has $T$ equal time intervals
 and $\mathcal{T}=\{1,2, \ldots, T\}$ denotes an index set of the time interval. For the $i$-th vehicle in time slot $t$, the generated computational task is denoted as $\mathcal{D}_{i}(t)=\left\{\varphi_{i}^{d a}(t), \varphi_{i}^{c o}(t)\right\}$, where $\varphi_{i}^{d a}(t)$ and $\varphi_{i}^{c o}(t)$ indicate the size of the computational task and the required computational resources to complete the task, respectively. We consider the case of binary offloading, where each vehicle chooses to locally process the task onboard, or offload the task to either a RSU or the MBS.

\vspace{-2pt}
\subsection{Local Computing Model}

When the task $\mathcal{D}_{i}(t)$ is executed locally, the latency only includes computation latency. The entire delay can be expressed as follows
\begin{equation}
	La_{i, l}^{\mathrm{loc}}(t)=\frac{\varphi_{i}^{c o}(t)}{F_{i}},
\end{equation}
where $F_i$ represents the computational resources of  vehicle $i$ itself.

\vspace{-2pt}
\subsection{RSU Computing Model}
The latency when task $\mathcal{D}_{i}(t)$ is offloaded to the $r$-th RSU includes computation and communication latencies. The communication transmission rate is given by
\begin{equation}
	\label{local}
	\operatorname{tr}_{i, r}^{RSU}(t)=B_{RSU} \cdot \log _2\left(1+\frac{p_t \cdot  g_{i, r}(t)}{\sigma^2 \cdot s_{i, r}(t)^2}\right),
\end{equation}
where $B_{RSU}, p_t, \sigma^2,$ and $g_{i, r}(t)$ represent bandwidth, transmit power, signal noise, and channel gain, respectively. $s_{i, r}(t)$ is the distance between vehicle $i$ and RSU.

The entire delay is expressed as
\begin{equation}
	\label{RSU}
	La_{i, r}^{RSU}(t)=\frac{\varphi_{i}^{co}(t)}{f_{r, i}(t)}+\frac{\varphi_{i}^{da}(t)}{tr_{i, r}^{RSU}(t)},
\end{equation}
where $f_{r, i}(t)$ represents the computational resources allocated to vehicle $i$ by the $r$-th RSU at time slot $t$. It is calculated as $f_{r, i}(t) = \frac{\varphi_{i}^{co}(t)}{\zeta_r(t)} \cdot F_R$, with $\zeta_r(t)$ being the total computational resources served by the $r$-th RSU.

\vspace{-2pt}
\subsection{MBS Computing Model}
The latency when task $\mathcal{D}_{i}(t)$ is offloaded to the MBS consists of computation and communication latencies. The communication transmission rate is given by
\begin{equation}
	\operatorname{tr}_{i, MBS}^{MBS}(t)=B_{MBS} \cdot \log _2\left(1+\frac{p_t \cdot  g_{i, MBS}(t)}{\sigma^2 \cdot s_{i, MBS}(t)^2}\right),
\end{equation}
where $B_{MBS}, p_t, \sigma^2,$ and $g_{i, MBS}(t)$ represent bandwidth, transmit power, signal noise, and channel gain, respectively. $s_{i, MBS}(t)$ is the distance between vehicle $i$ and the MBS.

The entire delay is expressed as
\begin{equation}
	\label{cloud}
	La_{i, MBS}^{MBS}(t)=\frac{\varphi_{i}^{co}(t)}{f_{MBS, i}(t)}+\frac{\varphi_{i}^{da}(t)}{tr_{i, MBS}^{MBS}(t)},
\end{equation}
where $f_{MBS, i}(t)$ represents the computational resources allocated to vehicle $i$ by the MBS at time slot $t$. It is calculated as $f_{MBS, i}(t) = \frac{\varphi_{i}^{co}(t)}{\zeta_{MBS}(t)} \cdot F_{MBS}$, with $\zeta_{MBS}(t)$ being the total computational resources served by the MBS.

\vspace{-2pt}
\subsection{Problem Formulation}
We investigate the optimization computation offloading problem in the proposed cybertwin-enabled IoV system. Our goal is to minimize the latency by tuning the offloading choices of vehicles, which is curial for computation-intensive real-time applications, such as automatic driving. The offloading choice of vehicle $i$ at time slot $t$ is expressed as $\psi_i(t) = \left[\psi_{i}^{local}(t), \psi_{i}^1(t), \ldots, \psi_{i}^r(t), \ldots, \psi_{i}^{R}(t), \psi_{i}^{MBS}(t) \right]$, Summarizing the above equations (\ref{local}), (\ref{RSU}), (\ref{cloud}), the total delay can be expressed as follows
\begin{equation}
	\begin{aligned}
		La_i(t) &= \psi_{i}^{local} \cdot La_{i, l}^{\mathrm{loc}}(t) + \sum_{\forall r \in \mathcal{R}} \psi_{i}^r \cdot La_{i, r}^{R S U}(t) \\
		&+ \psi_{i}^{MBS} \cdot La_{i, MBS}^{MBS}(t).
	\end{aligned}
\end{equation}

Our optimization problem can be expressed as follows
\begin{equation}
	\begin{array}{l}
		\min  \quad \sum_{t = 0}^{T} \sum_{i = 0}^{I} La_i(t)  \\
		\text { s.t. } \quad \sum\psi_i(t) = 1, \forall i \in \mathcal{I},
	\end{array}
\end{equation}
where constraint guarantees that each vehicle chooses one offloading option in each time slot.

\section{Knowledge-Driven Multi-Agent Reinforcement Learning Approach}
\label{sec:algorithm}
Traditional optimization methods can be used for problem-solving. However, the computational complexity explodes, especially with more vehicles, prompting the adoption of a reinforcement learning (RL) approach. RL offers an efficient way to obtain optimal strategies by allowing agents to learn from their interactions with the environment.

As the number of vehicles grows significantly, centralized DRL algorithms face challenges due to the exponential growth of the action space. This hinders their practicality in large-scale scenarios. Moreover, fully distributed DRL algorithms struggle with communication issues among multiple agents, which can lead to difficulties in achieving convergence. To overcome these limitations, the concept of CTDE in multi-agent reinforcement learning emerges as a potential solution.

In the context of CTDE, DRL requires information from all agents' observations during the training process. This involves concatenating observations and actions in a specific order that connects all agents. However, traditional DNN are sensitive to permutation changes, meaning different orders may produce different outputs, which contradicts the reality of the problem \cite{liu2020pic}. Consequently, this permutation sensitivity can lead to inefficient neural network training and hinder model scalability. To address this issue, in this paper, we introduce GNN to leverage communication topology and permutation invariance and propose knowledge-driven multi-agent reinforcement learning (KMARL). 
\subsection{Problem Transformation}
\subsubsection{Observation State}
Cybertwin $\widehat{i}$ collects localized state information before the start of each time slot including, vehicle $i$'s coordinates, vehicle $i$'s task size and vehicle $i$'s computational resource requirements. As a result, the local observations are as follows
$$O_i(t) = \left[ loca_i(t), \varphi_{i}^{d a}(t), \varphi_{i}^{c o}(t)\right].$$

\subsubsection{Action State}
Cybertwin $\widehat{i}$ can decide where to execute the task between locally, RSUs, and MBS. The whole action space is a $(R + 2)$ column vector, where $\psi_{i}^{local} = 1$ indicates that the task is executed locally, $\psi_{i}^{r} = 1$ indicates that the task is offloaded to the $r$-th RSU, and $\psi_{i}^{MBS} = 1$ instructs the task to be offloaded to the MBS. The action space can be represented as follows
$$A_i(t)=\left[\psi_{i}^{local}, \psi_{i}^1, \ldots, \psi_{i}^r, \ldots, \psi_{i}^{R}, \psi_{i}^{MBS}\right].$$

\subsubsection{Reward Function}
The behavior of agents is reward-driven, which means that the reward function is a key step in the training phase. The goal of this paper is to minimize the task latency by adjusting the vehicle computation offloading selection. And in this cooperative task, all the agents share the team reward. We then construct a reward function based on the objective function
\begin{equation}
	\mathcal{R}(t)= -\frac{1}{I}\sum_{\forall i \in \mathcal{I}}\left[\eta_{i, t} + La_i(t)\right].
\end{equation}
The $\eta_{i, t}$ indicates the failure penalty, which correlates with the relative magnitudes of the actual time delay and the locally computed time delay.

\subsection{The Design of KMARL Approach}
\subsubsection{The Basic Concept of KMARL}
\begin{figure}[htbp]
	\centering
	\includegraphics[width=.49\textwidth]{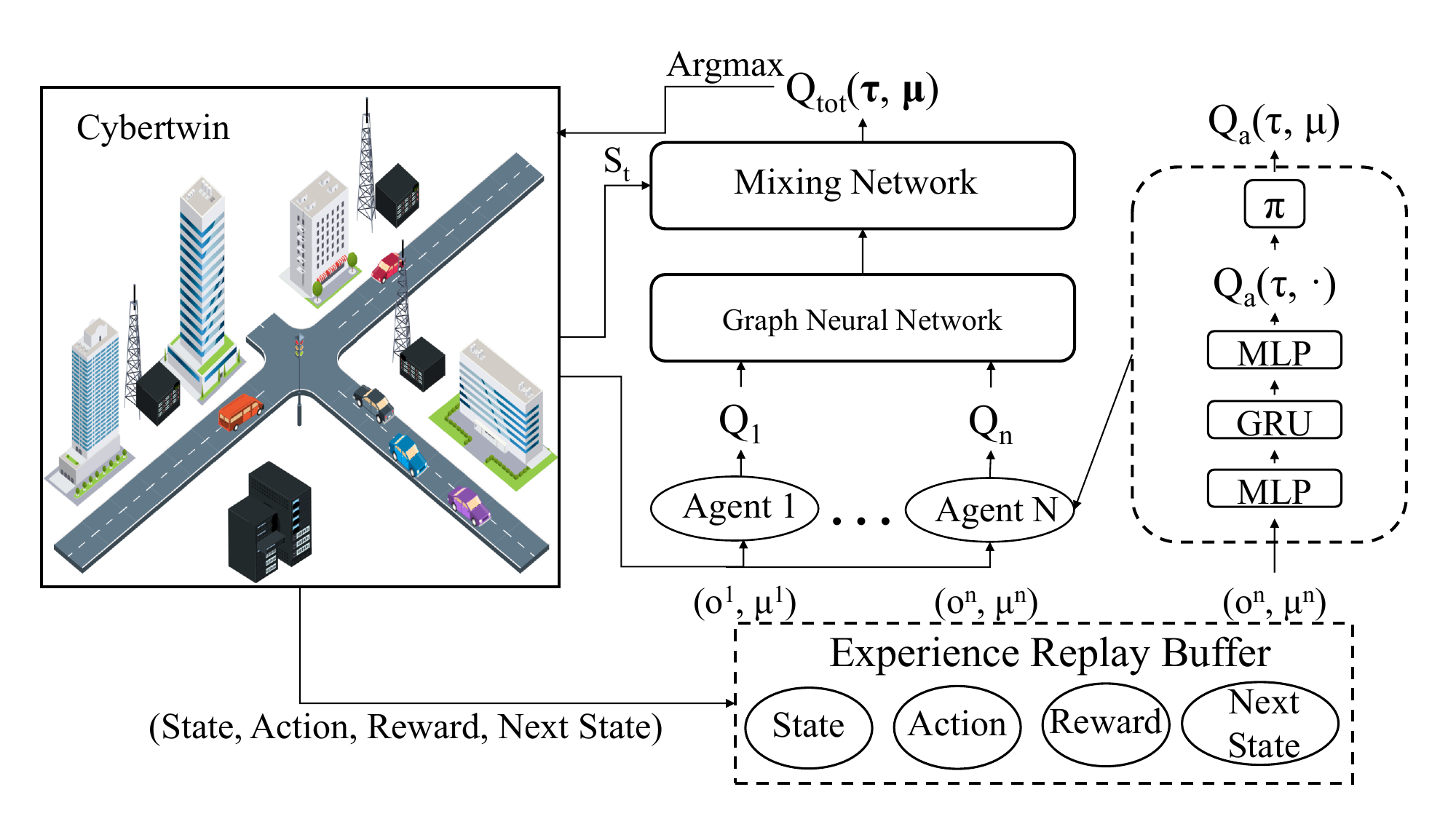} 	
	\caption{The network architecture of KMARL.}		
	\label{RL}							
\end{figure}

In this subsection, we introduce the KMARL network architecture, as shown in Fig. \ref{RL}. KMARL is a QMIX-based multi-agent reinforcement learning algorithm incorporating communication topology and permutation invariance. In this method, cybertwin is deployed on an RSU cluster, and the agents are trained to learn offloading decisions. During the policy update process, each cybertwin observation corresponds to objects in the actual network, and the offloading strategy is determined based on the observations.

Here we will first discuss QMIX, with the GNN section \ref{gcn} following in more detail. QMIX is designed to learn the joint action-value function $Q_{tot}$ with access to global state information. Compared to fully distributed algorithms, distributed agents can choose more reasonable independent actions based on local observations. The network architecture of QMIX algorithm primarily consists of two key components: local value function networks and centralized mixing network.

The architecture of the local value function networks is depicted on the right side of Fig. \ref{RL}, and it is responsible for computing the action values for individual agents. The centralized mixing network takes the action values of all agents as inputs and combines the entire state space as weighted parameters for each action value. The constraint between the total action value $Q_{tot}$ and any individual action value $Q_a$ can be represented as:
$$\frac{\partial Q_{tot}}{\partial Q_a} \geq 0 \quad \forall a \in A.$$

The QMIX algorithm utilizes a loss function based on the bellman equation to train the centralized mixing network and local value function networks. The target value is obtained by adding the actual reward under the current observed state to the maximum value of the global value function under the next state, represented as $y^{tot}=r+\gamma \max_{u'} Q_{tot}(\boldsymbol{\tau}', \boldsymbol{u}', s'; \theta^-)$, where $\theta^-$ represents the parameters of the target network, $\tau$ denotes the agent's past history, $u$ is the joint action output of all agents, and $s$ represents the global state. The global value function is directly generated by the centralized mixing network. A fixed target network is used in the target value computation to reduce instability during the training process. The overall loss function can be expressed as:
\begin{equation}
	\mathcal{L}(\theta)=\sum_{i=1}^I\left[\left(y_i^{t o t}-Q_{t o t}(\boldsymbol{\tau}, \boldsymbol{u}, s ; \theta)\right)^2\right].
\end{equation}

The network parameters are updated through backpropagation in KMARL. Moreover, our algorithm adopts a value-based multi-agent reinforcement learning approach, which is different from policy-based multi-agent learning like MADDPG. It is well-suited for binary computation offloading and does not require further regularization of the input state space, among other benefits.

\subsubsection{Graph Neural Network}
\label{gcn}
GNN is placed before the centralized mixing network, where all vehicles are modeled as a graph. Each node represents an agent, and each edge represents the relationship between nodes. Based on this, agents achieve the aggregation of neighboring nodes through the convolutional layer, and finally realize the permutation invariance. We denote the $L$ convolutional layers as $\left\{f_{\mathrm{GNN}}^{(1)}, \ldots, f_{\mathrm{GNN}}^{(L)}\right\}$. The graph convolutional layer takes the node representations and the adjacency matrix of the graph as input and computes new representations for each node. The updates between different convolutional layers can be represented as:
\begin{equation*}
	\begin{aligned}
		\mathbf{h}^{(l)} & =f_{\mathrm{GNN}}^{(l)}\left(\mathbf{h}^{(l-1)}\right) \\
		& =\sigma\left(\frac{1}{N} A_{\text {adj }} \mathbf{h}^{(l-1)} W_{\text {other }}^{(l)}+\operatorname{diag}(1)_N \mathbf{h}^{(l-1)} W_{\text {self }}^{(l)}\right),
	\end{aligned}
\end{equation*}
where $A_{adj}$ represents the adjacency matrix among the vehicles, with elements $A_{ij}$ indicating whether there is a connection or communication between the i-th and j-th vehicles. $W_{\text{other}}^{(l)}$ and $W_{\text{self}}^{(l)}$ are trainable weight matrices between layers, and $\sigma$ is a nonlinear activation function.

Next, a pooling layer is applied to the output of the $L$-th convolutional layer $\mathbf{h}^{(L)}$. Both max pooling and average pooling satisfy permutation invariance since summation and element-wise maximization are commutative operations. Through the $L$ convolutional layers and the final pooling layer, message passing between vehicles is achieved, incorporating the permutation invariance inherent in GNN.
\section{Performance Evaluation}
\label{sec:simu}
\subsection{Simulation Settings}
We consider that the vehicles maintain a constant speed without any turning or reversing to simplify the research problem. The size of the task, $\varphi_{i}^{d a}(t)$, is in the range of [1, 1.5, 2] Mbits. The required computational resources for the task, $\varphi_{i}^{c o}(t)$, are proportional to the task size, and we introduce a linear constant, $\rho$, generated randomly between 100 and 200, such that $\varphi_{i}^{c o}(t) = \rho \cdot \varphi_{i}^{d a}(t)$. The transmit power of the vehicles is set to 20 dBm. The RMSProp optimizer is used to update the neural networks. Some important parameters are listed in Table \ref{tab:para}.

\begin{table}
	\centering
	\caption{EXPERIMENTAL PARAMETER SETTING}
	\label{tab:para}
	\begin{tabular}{ll}
		\toprule
		Parameter & Value  \\
		\midrule
		CPU cycle of vehicle ($F_i$) & $5 \cdot 10^{5}Hz$  \\
		CPU cycle of RSU ($F_R$)& $6 \cdot 10^{6}Hz$   \\
		CPU cycle of MBS ($F_{MBS}$) & $10^{7}Hz$  \\
		Bandwidth of vehicle to RSU channel ($B_{RSU}$) & $2 \cdot 10^{8}Hz$ \\
		Bandwidth of vehicle to MBS channel ($B_{MBS}$) & $2 \cdot 10^{7}Hz$ \\
		Experience replay buffer size & 2000 \\
            Discount factor & 0.9 \\
            Learning rate & $10^{-4}$ \\
            Batch size & 64 \\
		\bottomrule
	\end{tabular}
\end{table}

To demonstrate the superiority of the agent trained by our proposed method in terms of training and decision efficiency, we compare the KMARL method with the following benchmark methods, i.e.,
\begin{itemize}
	\item  Value Decomposition Networks (VDN): VDN has the property of permutation invariance, but does not incorporate information about the state of the environment. 
	\item QMIX: QMIX incorporates state information and does not have the property of permutation invariance.
\end{itemize}
\subsection{Performance Comparison}
\begin{figure}[htbp] 
	\centering
	\includegraphics[width=.49\textwidth]{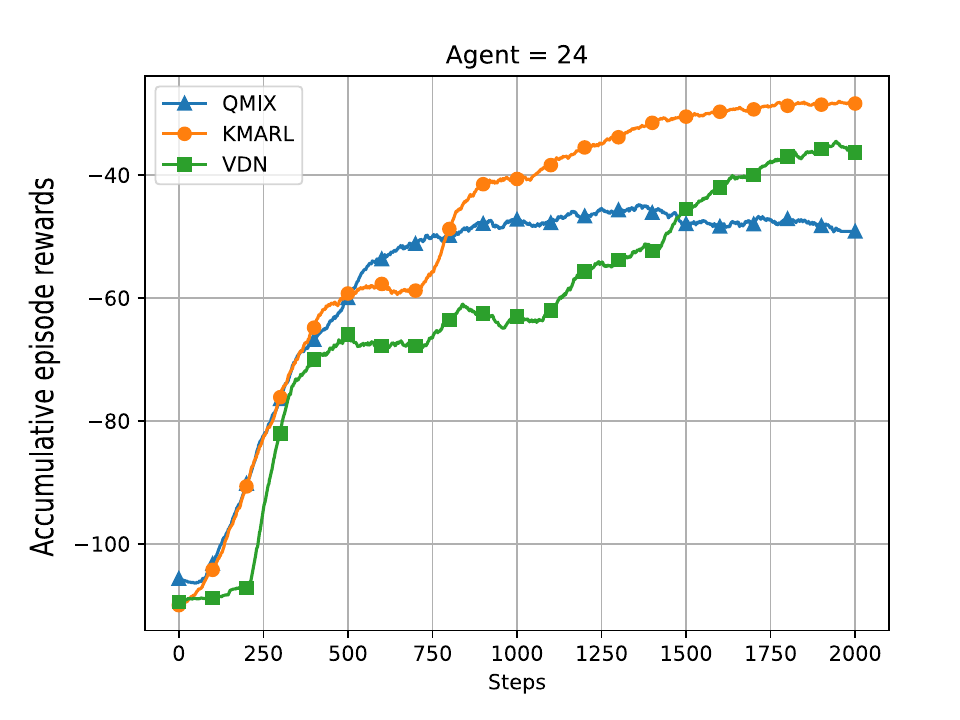} 	
	\caption{Accumulative episode reward.}		
	\label{result}							
\end{figure}

Fig. \ref{result} illustrates the convergence performance when the number of vehicles is 24. We can obtain that KMARL algorithm achieves the highest reward, followed by VDN and QMIX. This is because KMARL incorporates communication topology, permutation invariance and state information of the environment. Meanwhile, VDN achieves a higher reward than QMIX illustrating the improvement in reward due to permutation invariance.

\begin{figure}[htbp]
	\centering
	\includegraphics[width=.49\textwidth]{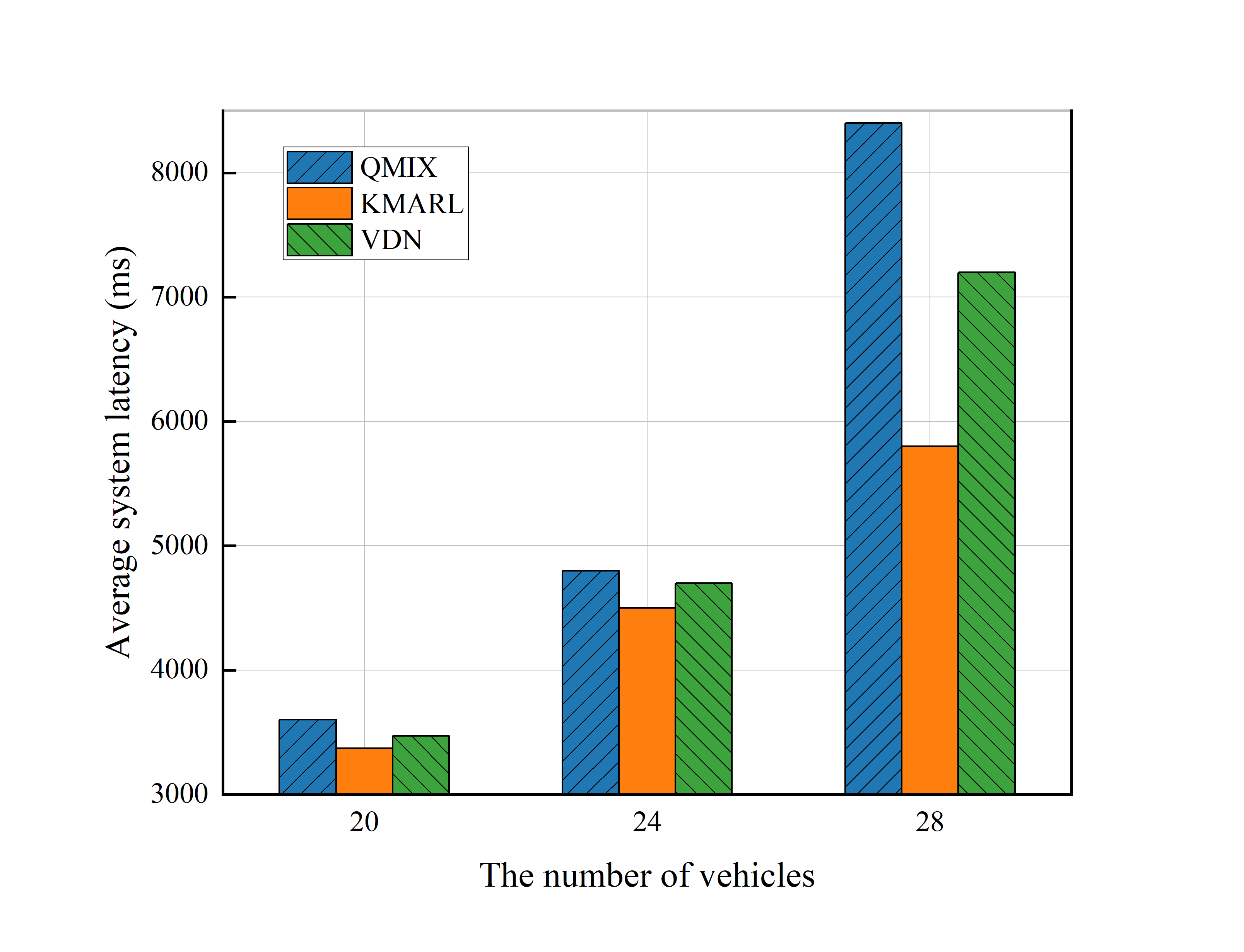} 	
	\caption{ Average system latency v.s. number of vehicles.}		
	\label{result-number}							
\end{figure}

Fig. \ref{result-number} shows the comparison of average system latency while the different numbers of vehicles. It can been seen that the value of average system latency increases linearly as the number of vehicles grows. It is because more vehicles can generate more computation tasks in IoV systems. It is worth noting that the KMARL algorithm has some performance improvement for the delay of the IoV system when the number of vehicles is 20 or 24; and the KMARL algorithm becomes especially significant for the delay performance improvement of the system when the number of vehicles is 28. This is because our method reasonably incorporates communication topology and permutation invariance by modeling the nodes as graphs, and incorporates the environmental state information. Overall, we can get that by incorporating permutation invariance, the algorithm can achieve a high reward value improvement, and our KAMRL algorithm exhibits improved scalability.

\section{Conclusion}
\label{sec:conc}
In this paper, we have investigated the knowledge-driven computation offloading problem in cybertwin-enabled IoV to minimize the latency of the task. In the proposed scheme, we have designed the cybertwin for each vehicle to help vehicles manage information and make offloading decisions. Besides,  a KMARL method based on QMIX algorithm has been proposed to train a decision-making agent and achieve the optimal offloading strategy.
Extensive experiments have been conducted, and results demonstrate that KMARL is well-performed than other comparison methods both in reward and model scalability. 
In future work, we will study the performance of computation offloading in more complex problems such as limited bandwidth.

\section*{Acknowledgment}
This work was supported in part by the National Key Research and Development Program of China under Grant 2020YFB1807700, in part by the National Natural Science Foundation of China (NSFC) under Grant 62201414 and 62071356, and in part by  the Fundamental Research Funds for the Central Universities under Grant ZYTS23175.

\small
\bibliographystyle{IEEEtran}
\bibliography{MyRefs}

\end{document}